\documentclass[letterpaper]{article} 
\usepackage{aaai24}  
\usepackage{times}  
\usepackage{helvet}  
\usepackage{courier}  
\usepackage[hyphens]{url}  
\usepackage{graphicx} 
\usepackage{natbib}  
\usepackage{caption} 
\usepackage[linesnumbered,ruled,vlined]{algorithm2e}
\usepackage{algorithmic,float}
\usepackage{float}
\usepackage{booktabs}
\usepackage{multirow}
\usepackage{multicol}
\usepackage{amssymb}
\usepackage{bbding}
\usepackage{amsmath}
\usepackage{changepage}

\usepackage{newfloat}
\usepackage{listings}
\DeclareCaptionStyle{ruled}{labelfont=normalfont,labelsep=colon,strut=off} 
\floatstyle{ruled}
\newfloat{listing}{tb}{lst}{}
\floatname{listing}{Listing}
\newcommand{\para}[1]{{\textbf{#1}}}
\newcommand{\method}[0]{\textbf{S$^3$A}}
\newcommand{\ie}[0]{\textit{i.e.}}
\newcommand{\eg}[0]{\textit{e.g.}}

\SetKwInput{KwInput}{Input}
\SetKwInput{KwOutput}{Output} 

\title{S3A: Towards Realistic Zero-Shot Classification via Self Structural Semantic Alignment}
\author{
    Sheng Zhang,\textsuperscript{\rm 1}
    Muzammal Naseer,\textsuperscript{\rm 1} 
    Guangyi Chen,\textsuperscript{\rm 1, \rm 2} 
    Zhiqiang Shen,\textsuperscript{\rm 1} \\ 
    Salman Khan,\textsuperscript{\rm 1, \rm 3}
    Kun Zhang,\textsuperscript{\rm 1, \rm 2} 
    Fahad Khan \textsuperscript{\rm 1, \rm 4} \\
}
\affiliations{
    \textsuperscript{\rm 1}Mohamed bin Zayed University of Artificial Intelligence,\\
    \textsuperscript{\rm 2}Carnegie Mellon University,\\
    \textsuperscript{\rm 3}Australian National University,\\
    \textsuperscript{\rm 4}Link\"{o}ping University\\
    \{firstname.lastname\}@mbzuai.ac.ae
}

\usepackage{bibentry}

\begin{document}

\maketitle

\begin{abstract}
Large-scale pre-trained Vision Language Models (VLMs) have proven effective for zero-shot classification. Despite the success, most traditional VLMs-based methods are restricted by the assumption of partial source supervision or ideal target vocabularies, which rarely satisfy the open-world scenario. In this paper, we aim at a more challenging setting,  \textit{Realistic} Zero-Shot Classification, which assumes no annotation but instead a broad vocabulary. 
To address the new problem, we propose the Self Structural Semantic Alignment (\textbf{S\textsuperscript{3}A}) framework, which extracts the structural semantic information from unlabeled data while simultaneously self-learning. 
Our \textbf{S\textsuperscript{3}A} framework adopts a unique Cluster-Vote-Prompt-Realign (CVPR) algorithm, which iteratively groups unlabeled data to derive structural semantics for pseudo-supervision. 
Our CVPR algorithm includes iterative clustering on images, voting within each cluster to identify initial class candidates from the vocabulary, generating discriminative prompts with large language models to discern confusing candidates, and realigning images and the vocabulary as structural semantic alignment. 
Finally, we propose to self-train the CLIP image encoder with both individual and structural semantic alignment through a teacher-student learning strategy.
Our comprehensive experiments across various generic and fine-grained benchmarks demonstrate that the \textbf{S\textsuperscript{3}A} method substantially improves over existing VLMs-based approaches, achieving a more than $15\%$ accuracy improvement over CLIP on average. 
Our codes, models, and prompts are publicly released at https://github.com/sheng-eatamath/S3A. 
\end{abstract}

\section{Introduction}
In recent years, large-scale pre-trained Vision Language Models (VLMs) such as CLIP \cite{clip, open-clip}, ALIGN \cite{AlignAP}, and BLIP \cite{blip, blip-2} have garnered significant attention for their remarkable zero-shot generalization ability on multifarious downstream tasks, particularly in recognizing unseen categories~\cite{survey-vlm}. 
The common practice to leverage this ability is packing category names into a textual prompt (e.g., ``A photo of a \texttt{[CLS]}'') and aligning image embeddings with text embeddings of filled prompts in VLM joint embedding space for classification. 
To adapt pre-trained VLMs to downstream unseen data, existing prevailing methods~\cite{survey-ovl, ov-detr, ov-seg} usually assume the access to source labeled data~\cite{plot, maple, coop} (\eg, in zero-shot learning~\cite{must-13, must-14}), target label distribution (\eg, in unsupervised prompt tuning~\cite{prompt-ldp}), or an \textit{ideal vocabulary} that exactly matches the ground-truth label set or with very few open words (\eg, in open-vocabulary learning~\cite{survey-ovl, ov-detr, ov-seg}). 
However, this ideal vocabulary is unattainable without exhaustive annotation of all unseen data; whereas, human annotations are exorbitant and difficult to scale.
Therefore, both assumptions are restrictive and impractical in open-world scenarios with diverse and dynamic nature. \par 

In this paper, we embark on a journey towards \textit{Realistic} Zero-Shot Classification (RZSC), a more challenging yet practical problem compared with conventional zero-shot learning due to its realistic conditions.
Here, we term \textit{Realistic} as the realistic nature of RZSC which aims to recognize categories on unseen datasets without annotation and ideal vocabulary, but with a vast, comprehensive vocabulary with more than 20K category names encompassing all common classes~\cite{concept-gen, imagenet21k}. 
However, it is challenging since the vast vocabulary can lead to alignment confusion among fine-grained options; as we witness the consistent and dramatic CLIP~\cite{clip} performance drops and reduced neighborhood ranges in Fig.~\ref{exp:motivation}. \par 

To confront this challenge, we introduce the Self Structural Semantic Alignment (\textbf{S\textsuperscript{3}A}) framework, which iteratively discovers structural semantic alignment from unlabeled data for joint self-learning.
This is orchestrated through our unique Cluster-Vote-Prompt-Realign (CVPR) algorithm, a principled process comprising four key steps: \texttt{(1)} \textbf{Clustering} unearths inherent grouping structures of image embeddings, producing meaningful image semantics. 
\texttt{(2)} \textbf{Voting} associates each cluster with initial category candidates, representing potential structural semantic alignments.
These two steps can be executed iteratively to obtain more reliable candidates.
\texttt{(3)} \textbf{Prompting} leverages the power of large language models (LLMs) to discern nuanced candidates by augmenting prompts with discriminative attributes.
\texttt{(4)} \textbf{Re-alignment} represents calibrating the cluster-vocabulary alignment with LLM-augmented prompts as pseudo structural semantic alignment labels.
Incorporating our CVPR algorithm, our \textbf{S\textsuperscript{3}A} framework self-trains a student model based on derived individual and structural semantic alignment labels from a stable teacher. Simultaneously, the teacher is updated by student weights to produce more reliable pseudo semantic alignments. \par 

We extensively evaluate our \textbf{S\textsuperscript{3}A} framework across multiple setups, spanning various generic and fine-grained benchmarks. 
The results show that \textbf{S\textsuperscript{3}A} not only consistently outperforms previous adapted State-of-The-Arts (SOTAs) under the RZSC setting on all benchmarks, but excels in out-of-vocabulary evaluation, where category names can fall outside the \textbf{S\textsuperscript{3}A} vocabulary.
Comprehensive evaluations evidence our \textbf{S\textsuperscript{3}A} framework effectively addressed this realistic challenging problem. \par 

\noindent{\textbf{Our contributions}} include: \texttt{(1)}
We propose a Self Structural Semantic Alignment (\textbf{S$^3$A}) framework, to address the challenging \textit{Realistic Zero-Shot Classification} problem, which 
jointly extracts and self-learns on the individual and structural semantic alignment.
\texttt{(2)} We propose a Cluster-Vote-Prompt-Realign algorithm to reliably derive reliable structural semantic alignments between images and the large vocabulary.
\texttt{(3)} \textbf{S$^3$A} achieves SOTA performance on various generic and fine-grained benchmarks, remarkably boosting CLIP by over $15\%$ accuracy, and even in the out-of-vocabulary scenarios.\par

\section{Related Work}
\begin{table}[]
    \centering
    \fontsize{9pt}{9pt}
    \selectfont
    \begin{tabular}{c|ccc}
    \toprule
        Setting & Vocab. & Anno. & Train \\
        \midrule \midrule
        Zero-Shot Transfer & $\mathcal{Y}_{tgt}$ & \XSolidBrush & $\mathcal{Y}_{tgt}$ \\
        Zero-Shot Classification & $\mathcal{Y}_{tgt}$ & \Checkmark & $\mathcal{Y}_{base}$ \\
        Open-Vocabulary Learning & $<$2K & \Checkmark & $\mathcal{Y}_{base}$ \\
        Unsupervised Fine-tuning & $\mathcal{Y}_{tgt}$ & \XSolidBrush & $\mathcal{Y}_{tgt}$ \\
        \midrule
        RZSC & $>$20K & \XSolidBrush & $\mathcal{Y}_{tgt}$ \\
         \bottomrule
    \end{tabular}
     \caption{Our realistic zero-shot classification and other related settings. Here, following~\cite{survey-ovl}, we denote $\mathcal{Y}_{base}$ and $\mathcal{Y}_{tgt}$ as sets of base training classes and target testing classes, which satisfies $\mathcal{Y}_{base} \bigcap \mathcal{Y}_{tgt}=\phi$. The learning goal of all settings is to recognize $\mathcal{Y}_{tgt}$ in test data. }
    \label{tab:setting}
\end{table}

\para{Zero-Shot Learning/Open-Vocabulary Learning with VLMs.} 
Traditional \textit{(Generalized) Zero-Shot Classification} (ZSC) aims to categorize novel classes in unseen test data with training on annotated base/seen classes with or without unlabeled target novel classes~\cite{survey-zsl-a, survey-gzsl}.
However, they usually assume auxiliary semantic information of both seen and unseen classes, \eg, category attributes~\cite{survey-zsl-a-80}, knowledge graph~\cite{survey-zsl-a-3}, textual keywords~\cite{survey-zsl-a-8, survey-zsl-a-13}.
Recently, large-scale pre-trained VLMs have been introduced to alleviate these assumptions~\cite{align, clip, survey-vlm}. 
Furthermore, \textit{Open-Vocabulary Learning} (OVL)~\cite{survey-ovl,zegclip, maskclip, diffusion-zsseg} aims to train the models with some annotated data, \ie, base classes, or large-scale image-text pairs, and to test them on target novel classes~\cite{diffusion-ovseg, unsup-zst}. 
Our RZSC setting differs from conventional ZSC and OVL in not requiring any labeled training data, 
and not assuming an \textit{ideal vocabulary} with a ground-truth target label set or one with few open words~\cite{survey-ovl, diffusion-ovseg, diffusion-zsseg}. \par

\begin{figure}[!t]
\centering
  \begin{minipage}{.48\linewidth}
  	\centering
    \includegraphics[ width=\linewidth,keepaspectratio]{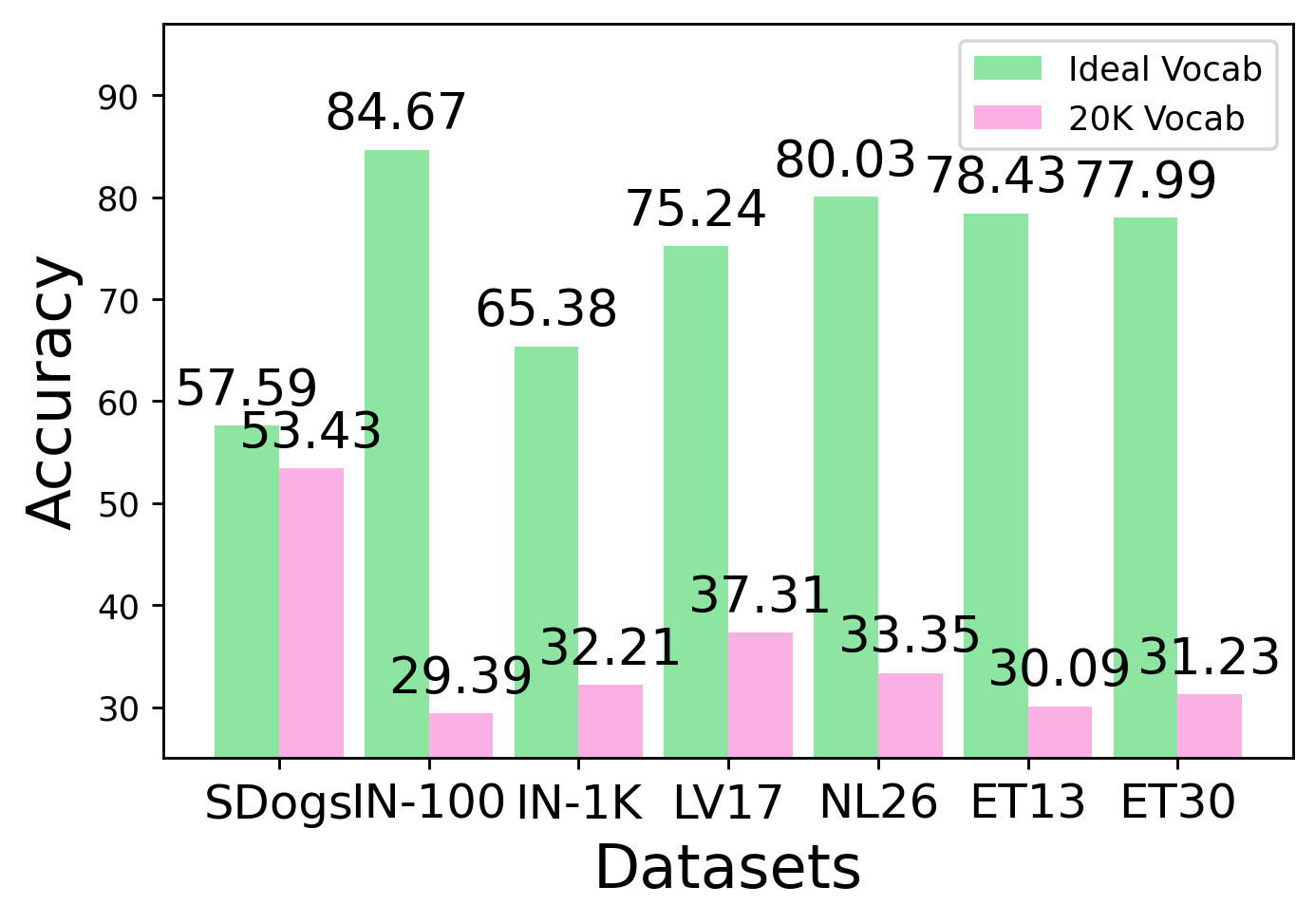}
  \end{minipage}
  \begin{minipage}{.48\linewidth}
  	\centering
    \includegraphics[width=\linewidth, keepaspectratio]{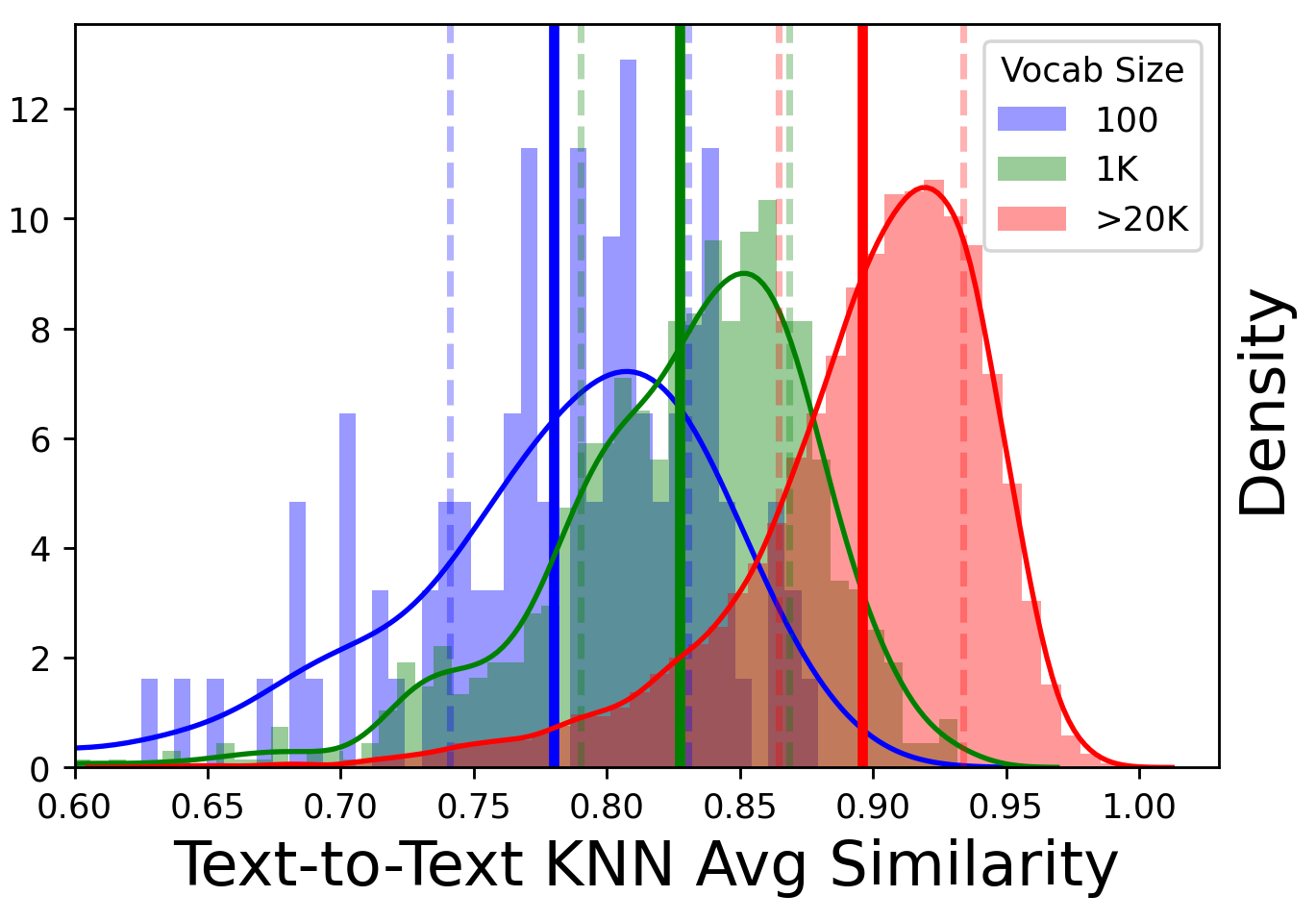}
  \end{minipage}
    \caption{\textbf{(a)} Performance comparison between CLIP w/ an ideal vocabulary (Green) and w/ a large vocabulary of 20K categories (Pink). \textbf{(b)} Distribution plot of text-to-text average 3-Nearest Neighbors cosine similarity of each text embedding for three types of vocabulary: with ImageNet-100, ImageNet-1K, and 20K category names.}
\label{exp:motivation}
\end{figure}

\para{Zero-Shot Transfer/Unsupervised Fine-tuning of VLMs.} 
Both \textit{Zero-Shot Transfer} (ZST) and \textit{Unsupervised Fine-tuning} (UF) assume no annotations of target datasets, which are essentially visual concept discovery problems~\cite{gcd, simgcd, promptcal} with vocabulary prior.
ZST~\cite{clip, open-clip} directly uses the pre-trained VLMs for zero-shot prediction without fine-tuning.
UF further transductively adapts pre-trained models with task-specific training, \eg, with self-training or prompt tuning~\cite{must, prompt-ldp, unsup-zst}.
However, both ZST\&UF assume known ground-truth target label sets or distribution~\cite{prompt-ldp, must}.
In this paper, we aim to alleviate the reliance on these assumptions and propose a new setting, RZSC. 
Besides, an extended ZST work, SCD~\cite{scd}, iteratively refines CLIP zero-shot inference predictions on a WordNet vocabulary~\cite{wordnet} with a heuristic iterative clustering algorithm. However, they have limited adaptability~\cite{must, prompt-ldp} and a mismatched linguistic vocabulary. \par  

\para{Discussion on Zero-Shot Settings.} 
Here, we summarize the main differences between our RZSC setting and others in Table~\ref{tab:setting}.
Previous related settings adopt restrictive assumptions including an ideal vocabulary, the target label distribution, and labeled base classes.
By contrast, our RZSC aims to learn to categorize an unlabeled dataset with a huge vocabulary based on a large visual taxonomy with over 20K classes.
An expanded vocabulary presents significant challenges for RZSC problem, as evidenced by the consistent and substantial CLIP performance drop (Fig.~\ref{exp:motivation}a) on all datasets when the vocabulary scales up. The primary challenge arises from increased confusing open words, complicating fine-grained category discrimination for pre-trained VLMs. 
As displayed in Fig.~\ref{exp:motivation}b, the averaged cosine similarity between a query text embedding and its 3-nearest text neighbors grows with the vocabulary size. Additional related works are included in Appendix B. \par

\begin{figure}[tp]
    \centering
    \includegraphics[width=\linewidth]{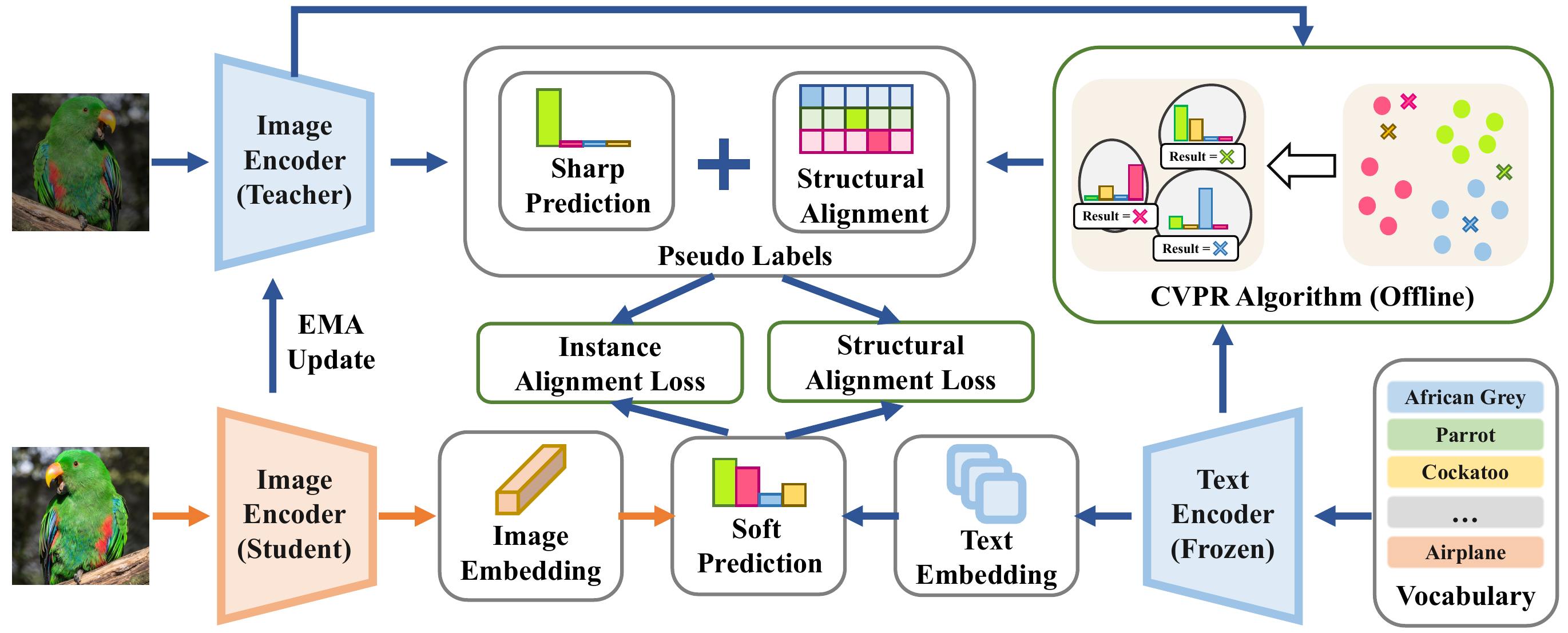}
    \caption{Illustration of our Self Structural Semantic Alignment (\method) framework, which fine-tunes pre-trained CLIP encoder with a teacher-student architecture. The teacher is updated by the student in an exponentially moving average manner. The student is guided by on-the-fly one-hot instance alignment predicted by the teacher, and self-trains with structural semantic alignment labels derived by our per-epoch CVPR algorithm on all teacher image embeddings.}
    \label{fig:sssa}
\end{figure}

\section{Methodology}

\subsection{Problem: Realistic Zero-Shot Classification}
Existing methods that adapt pre-trained VLMs to unseen data usually rely on specific knowledge of target datasets, such as prior distributions or an ideal vocabulary. These conditions are often challenging to fulfill in real-world environments. In this paper, we explore a more practical task, Realistic Zero-Shot Classification, abbreviated as RZSC. \par 

RZSC is formally defined as follows: Consider an unlabeled dataset $\mathcal{D}_u=\{(\mathbf{x}_i, y_i)\}_{i=1}^{N} \subset \mathcal{X}\times \mathcal{Y}$ with $N$ images, where $\mathcal{Y}$ is the underlying category set, and a pre-trained VLM such as CLIP, equipped with image and text encoders $f_I$ and $f_T$, respectively. 
Then, we assume no information of $\mathcal{Y}$ and instead with a comprehensive vocabulary that contains more than 20,000 distinct category names, \ie, $|\mathcal{Y}| \ll |\mathcal{W}|$. We build our vocabulary from all visual categories from ImageNet1K~\cite{imagenet1k} and ImageNet21K~\cite{imagenet21k} datasets since they are annotated with expert taxonomic knowledge~\cite{wordnet} and encompasses most commonly-seen visual categories in the real world.
The goal of the RZSC task is to adapt the pre-trained VLM, \ie, ${f_I, f_T}$ to predict the correct category of an unseen dataset:
\begin{equation}
    \hat{y_i} = \arg \max_{w_j \in \mathcal{W}}{\mathbf{z_i} \cdot \mathbf{h}_j},
    \label{eq:zst}
\end{equation}
where $\mathbf{z}_i=f_I(\mathbf{x}_i)$ denotes the image embedding while text embedding $\mathbf{h}_j=f_T(w_j)$ are obtained with a text prompt, \eg, ``a photo of a \{category\}'', and the category name $w_j$. Here, we denote $\cdot$ as cosine similarity. \par

\subsection{Overview: Self Structural Semantic Alignment}
RZSC presents a more formidable challenge than previous tasks, primarily owing to the absence of label information and an increased vocabulary size. As illustrated in part (a) of Fig.~\ref{exp:motivation}, the performance of CLIP declines sharply as the vocabulary size increases. This decline can be attributed to the inclusion of confusing open words as hard negative samples, which introduces noise to pre-trained CLIP, hindering its ability to accurately identify image-category alignments. \par 

We are motivated to propose our Self Structural Semantic Alignment (\textbf{S\textsuperscript{3}A}) framework, which discovers the structural semantics through iterative self-alignment between visual images and textual vocabulary.
As shown in Fig.~\ref{fig:sssa}, our \textbf{S\textsuperscript{3}A} incorporates a Cluster-Vote-Prompt-Realign (CVPR) algorithm to derive structural semantics as alignment labels, and both models and pseudo alignments are iteratively refined during self-training.
Our CVPR algorithm and \textbf{S\textsuperscript{3}A} self-training procedure can achieve a synergistic effect: as training progresses in adapting representations, the teacher model can provide increasingly reliable pseudo alignments in subsequent iterations. Concurrently, the CVPR algorithm contributes structural semantics as a refined supervisory signal for subsequent self-training. We elaborate on all components in the sections that follow. \par 

\begin{figure*}[tp]
    \centering
    \includegraphics[width=0.9\linewidth]{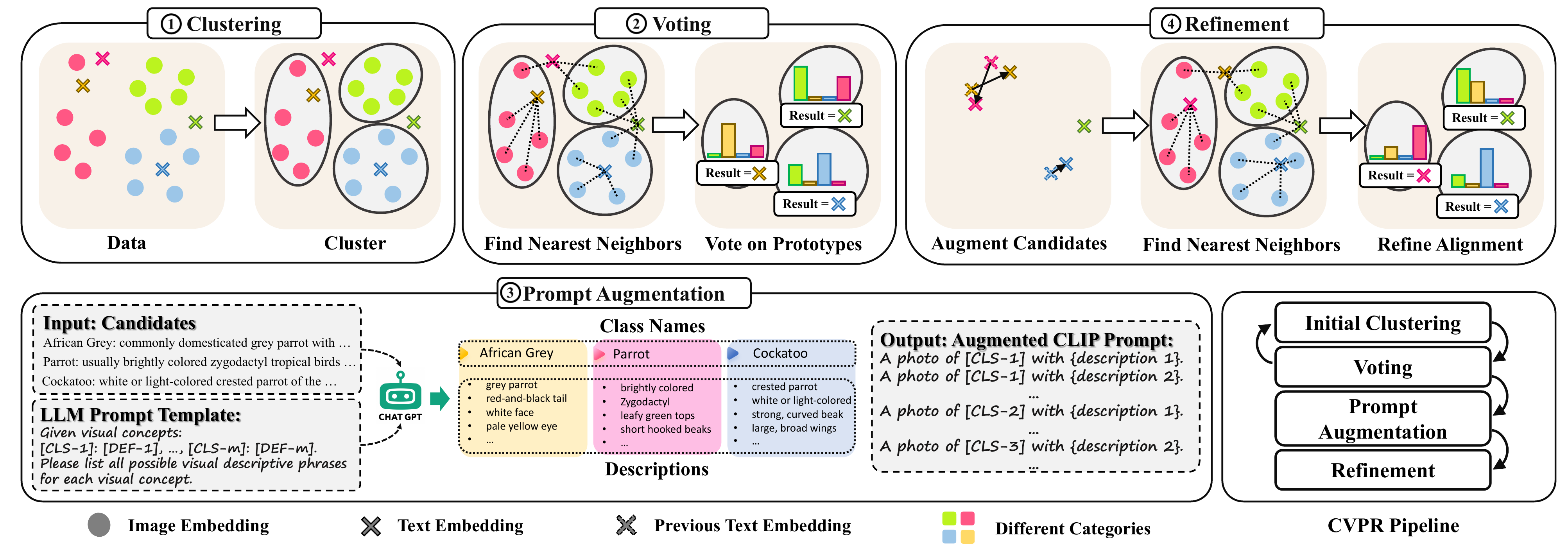}
    \caption{An illustrative toy example for our CVPR algorithm, comprising four steps: \texttt{(1)} We cluster all image embeddings. \texttt{(2)} We conduct 1-nearest neighbor voting on all text prototypes of the large vocabulary for each cluster.
    Since the results of the naive assignment in this step are susceptible to the noise of text embeddings, we generate cluster-wise candidate categories instead. \texttt{(3)} We augment CLIP text prompts with visual discriminative descriptions from the large language model to discern nuanced candidates. \texttt{(4)} With augmented prompts, the cluster-vocabulary alignment is calibrated and refined.}
    \label{fig:cvpr}
\end{figure*}

\subsection{Cluster-Vote-Prompt-Realign}
\label{sec:cvpr}
The Cluster-Vote-Prompt-Realign (CVPR) algorithm lies at the heart of the \textbf{S\textsuperscript{3}A} framework, representing an innovative approach to uncovering structural semantics in data. As illustrated in Fig.~\ref{fig:cvpr}, the CVPR algorithm consists of four key stages, each contributing to the alignment and identification of structural relationships between visual images and textual vocabulary, including discovering semantic clusters, voting category names on large vocabulary, prompting LLM to discriminate the nuanced candidates, and refine the cluster-vocabulary alignment. Each step is explained in detail in the subsequent paragraphs.
Below we delineate these stages and their functions within the algorithm.

\para{Clustering.} Based on existing evidence~\cite{clip} and our observation, the pre-trained CLIP excels at grouping instances with the same or similar semantic labels in the image embedding space. We thus produce the pseudo supervision by semantic clustering and aligning the clusters with vocabulary. Specifically, given image embeddings $\bf{z}_i$ in $\mathcal{D}_u$, we apply KMeans~\cite{kmeans} to obtain the $K$ clusters, $\Gamma=\{\Gamma_k\}_{k=1}^K$, where $\Gamma_k$ denotes the $k$-th set of image embeddings. \par 

\para{Voting.} Given the semantic cluster results $\Gamma$, we compute a vocabulary voting distribution matrix $M \in \mathbf{R}^{K \times |\mathcal{W}|}$, where $M_{k,j}$ represents the normalized frequency of the prototype of category $w_j$ being the nearest neighbor to all instances in the $k$-th cluster. Specifically, it is computed as
\begin{equation}
    M_{k,j} = 
    \frac{1}{K |\Gamma_k|} 
    \sum_{\mathbf{z}\in \Gamma_k}{\mathbb{I}(w_j = \arg \max_{w} \mathbf{z} \cdot \mathbf{h} ) }
    \label{eq:vote}
\end{equation}
where $\mathbb{I}$ is an indicator function, and $|\Gamma_k|$ denotes the size of the $k$-th cluster. 
$M$ is cluster-wise and vocabulary-wise normalized, with $||M||_1=1$.
Rather than naively assigning each cluster to the argmax prototype in the vocabulary, we keep the top-$m$ frequent words for each cluster as potential candidates which are treated equally. 
For each row $M_k=(M_{k,j})_{j=1}^{|\mathcal{W}|}$, we set all entries but the highest $m$ ones as 0. \par 

Nonetheless, the initial clustering and voting may introduce noise, leading to low-quality pseudo-labels. 
To mitigate this issue, we iteratively refine the previous clusters based on the current voting outcomes. 
In particular, we utilize the Hungarian matching~\cite{hungarian} for textual embeddings and clusters to align each cluster with a single prototype. Subsequently, we reassign the image embeddings, using these prototypes as the updated cluster centers~\cite{scd}.
Additional details are provided in our Appendix C. \par

\para{Prompting.} 
Through our empirical studies, we observed that CLIP representation struggles to differentiate nuanced candidates effectively. This observation spurred our efforts to refine the embeddings of textual candidates. We speculate that the challenge in distinguishing fine-grained options arises from the presence of noisy or ambiguous image-caption alignments during CLIP pre-training. \par 

To address this challenge, our approach is to enhance the conventional CLIP prompts by accentuating the subtle semantic differences. We achieve this by integrating auxiliary textual knowledge drawn from LLMs, which are effective in knowledge retrieval~\cite{gpt-3, gpt3-vqa}. Specifically, we feed $m$ candidate category words of the $k$-th cluster into a single LLM prompt template, each accompanied by their specific definition. Then, we add an instruction to the prompt to extract nuanced visual attributes of each category from the LLM. Our prompt template is structured as: \par 

\begin{adjustwidth}{10pt}{10pt}
\textbf{Prompt}: Given visual concepts: [CLS-1]: [DEF-1], ..., [CLS-m]: [DEF-m].
\end{adjustwidth}

\begin{adjustwidth}{10pt}{10pt}
\textbf{Goal}: To discriminate these visual concepts in a photo. Please list all possible visual descriptive phrases for each visual concept.
\end{adjustwidth} 

In this template, \texttt{[CLS]} represents the category name, and \texttt{[DEF]} stands for its definition from WordNet~\cite{wordnet}. 
The LLM then generates a list of distinctive attributes for each category, such as `red-and-black tail'. 
To avoid linguistic ambiguity arising from the polysemy phenomenon, we utilize all possible synset-definition pairs in WordNet~\cite{wordnet} for a single category as the input visual concepts for the LLM prompt.
Finally, each (category, attribute) pair is filled into a CLIP prompt for augmentation, \eg, ``A photo of a \{category\} with \{attribute\}.''. An ensemble of augmented text embeddings for each category name is constituted. \par

\para{Re-alignment.}
During the re-alignment phase, our goal is to enhance the structural semantic alignments in Eq.~\ref{eq:vote}.
The refined re-alignment matrix, $\tilde{M} \in \mathbf{R}^{K \times |\mathcal{W}|}$, is derived by casting votes on all augmented text embeddings generated in the previous prompting stage.
Specifically, the re-alignment probability between the $k$-th cluster and word $w_j$ is determined by the frequencies of augmented embeddings of the word $w_j$ in $\mathcal{A}_k$ being the top-3 nearest neighbors of $\mathbf{z}\in \Gamma_k$. We denote $\mathcal{A}_k$ as the set of augmented embeddings of all candidate category words of $\Gamma_k$. It can be formulated as: 
\begin{equation}
    \tilde{M}_{k,j} = \frac{\alpha_{w_j}}{3 K |\Gamma_k| }\sum_{\mathbf{z}\in \Gamma_k}{ \mathbb{I}\left(w_j \in \arg \mathop{\mathrm{top3}}_{w}(\mathbf{z}, \mathcal{A}_k)\right) }
    \label{eq:realign}
\end{equation}
where $\arg$ extracts the category name linked with the augmented text embedding in $\mathcal{A}_k$. To avoid the imbalance issue raised by varied numbers of augmented embeddings of different category names, we consider the weight factor $\alpha_{w_j}=\frac{1}{|\mathcal{A}_k(w_j)|}$, which uniformly distributes total mass $1$ to all augmented embeddings of $w_j$. Therefore, each row of $\tilde{M}$ sums to be $\frac{1}{K}$, and $||\tilde{M}||_1=1$.
We again employ the maximum Hungarian matching~\cite{hungarian} on a bipartite graph between clusters and category words, with cost matrix $\tilde{M}$. Consequently, the structural alignment is obtained, which enforces a one-to-one mapping between clusters and category names. \par

\subsection{Self Training with Semantic Alignment}
\label{method:sssa}
In this section, we present our S\textsuperscript{3}A self-training framework, as depicted in Fig.~\ref{fig:sssa}. 
The self-training process leverages both instance-wise and structural alignment pseudo labels which are derived by our CVPR algorithm with an exponentially moving averaged (EMA) teacher model~\cite{ema}.
Throughout this process, we adapt the CLIP image encoder to enhance its representation and fix the text encoder. \par 

\para{Structural Semantic Alignment.}
To incorporate the structural semantic alignments into online learning, one challenge needs to be addressed. 
Obtaining high-quality structural alignment pseudo-labels requires consistent model embeddings from the entire dataset, which is computationally costly; while determining the optimal execution interval of CVPR across datasets is challenging.
To mitigate these issues, we introduce a slowly updated EMA teacher model. It provides stably refined embeddings and executes the CVPR algorithm once per epoch to yield stable and reliable structural pseudo alignments, which then guides the self-training of the student model. \par 

We define the structural semantic alignment loss as the cross-entropy between the predictions of the student model and the pseudo structural alignments generated by the teacher model. Formally, this loss for the $i$-th instance can be expressed as:
\begin{equation}
   L_{str}(\mathbf{x}_i)= - \mathbf{\hat{p}}^T_T(i) \log \mathbf{p}_S(\mathbf{x}_i).
    \label{eq:str}
\end{equation}
In this equation, $\mathbf{\hat{p}}_T(i)$ represents the one-hot pseudo structural alignment for the $i$-th instance, which is inferred from the teacher CVPR results during the last epoch. 
On the other hand, $\mathbf{p}_S$ denotes the softmax prediction of the student model over the entire vocabulary, computed for the input $\mathbf{x}_i$. 
As a result, the sharpened pseudo labels can cluster images with the same semantics as well as align clusters. \par 

\para{Individual Semantic Alignment.} 
In addition to the structural semantic alignment loss, we also guide our model with instance-wise pseudo alignments, which are generated on-the-fly by the EMA teacher model. Without this guidance, our model would likely converge to suboptimal solutions rapidly. We formulate the individual semantic alignment loss for the $i$-th instance as follows:
\begin{equation}
    L_{in}(\mathbf{x}_i) = - \mathbb{I}(\mathbf{\tilde{p}}_T(\mathbf{x}_i) > \tau)  \mathbf{\tilde{p}}^T_T(\mathbf{x}_i) \log \mathbf{p}_S(\mathbf{x}_i).
    \label{eq:ins}
\end{equation}
In this equation, $\mathbf{\tilde{p}}_T$ represents the one-hot sharpened pseudo label produced by the teacher model at each iteration. The symbol $\tau$ denotes a confidence threshold, which ensures that the loss is computed only for samples for which the teacher model has a high level of confidence. \par 

To strike a balance between the structural and instance alignment losses, we introduce a weighted combination of both. In this way, individual alignment retains original instance alignment information, while structural alignment groups and aligns similar semantics.
Consequently, our total loss function for the $i$-th instance is formulated as:
\begin{equation}
    L(\mathbf{x}_i) = L_{str}(\mathbf{x}_i) + \gamma L_{in}(\mathbf{x}_i).
    \label{eq:total}
\end{equation}
Here, $\gamma$ represents a balancing factor that weights the contribution of the instance alignment loss relative to the structural alignment loss. This total loss is computed at each iteration, based on our CVPR algorithm which is executed once per epoch on the teacher model.

We present entire details of our \textbf{S\textsuperscript{3}A} training algorithm together with implementation details in our Appendix C. \par

\section{Experiments}
\label{exp}

\begin{table*}[htp]
    \centering
    \fontsize{9pt}{9pt}
    \selectfont
    \setlength{\tabcolsep}{4pt}
    \begin{tabular}{l|ccccccc|c}
    \toprule
        Methods & SDogs & IN100 & IN1K & LV17 & NL26 & ET13 & ET30 & Avg \\
         \midrule \midrule
         CLIP (Ideal) & 57.59/58.07 & 84.67/84.90 & 65.38/65.53 & 75.24/75.53 & 80.03/80.05 & 78.43/78.50 & 77.99/78.03 & 74.19/74.37 \\
         \midrule
         DINO+KMeans & -/45.99 & -/75.16 & -/55.27 & -/72.52 & -/62.81 & -/67.37 & -/64.69 & -/63.40 \\
         CLIP & 53.43/55.43 & 29.39/38.54 & 32.21/39.77 & 37.31/47.24 & 33.35/38.96 & 30.09/40.00 & 31.23/39.90 & 35.29/42.83 \\
         CLIP (Group) & 19.37/55.92 & 40.62/77.68 & 26.41/56.92 & 38.33/68.81 & 41.09/70.51 & 32.85/71.08 & 36.36/70.78 & 33.57/67.38 \\
         MUST & 57.20/60.61 & 33.37/52.56 & 28.97/37.00 & 31.71/49.35 & 35.30/48.68 & 38.46/58.25 & 33.41/47.08 & 36.92/50.50 \\
         SCD & 52.63/55.93 & 48.89/77.39 & 37.06/57.00 & 43.33/68.81 & 52.18/71.84 & 40.46/71.25 & 46.29/70.89 & 45.83/67.59 \\
         \midrule
         \textbf{S$^3$A} (Our) & \textbf{58.94/62.19} & \textbf{52.08/82.76} & \textbf{42.43/63.15} & \textbf{48.34/75.57} & \textbf{56.20/75.97} & \textbf{45.21/76.92} & \textbf{50.41/76.14} & \textbf{50.51/73.24} \\
         \bottomrule
    \end{tabular}
    \caption{Transductive evaluation on seven benchmarks. Top-1 classification accuracy scores (left of `/`) and clustering accuracy scores (right of `/`) are reported in percentage. We highlight the highest scores except for the upper bound.}
    \label{tab:main-merged}
\end{table*}

\begin{table}[t]
    \centering
    \fontsize{9pt}{9pt}
    \selectfont
    \setlength{\tabcolsep}{4pt}
    \begin{tabular}{c|ccc||cc|cc}
    \toprule
        \multirow{2}{*}{\#Row} & \multirow{2}{*}{Prompt} & \multirow{2}{*}{S.T.} & \multirow{2}{*}{$L_{str}$} & \multicolumn{2}{c}{ImageNet-100} & \multicolumn{2}{c}{Living17} \\
        & & & & Acc & Cluster & Acc & Cluster \\
    \midrule \midrule
        1 & \XSolidBrush & \XSolidBrush & \XSolidBrush & 48.89 & 77.39 & 43.33 & 68.81 \\
        2 & \Checkmark & \XSolidBrush & \XSolidBrush & 51.81 & 79.38 & 44.60 & 68.69 \\
        3 & \XSolidBrush & \Checkmark & \XSolidBrush & 46.23 & 81.49 & 46.28 & 74.60 \\
        4 & \XSolidBrush & \Checkmark & \Checkmark & 49.00 & 82.08 & 46.55 & 73.04 \\
        \midrule
        5 & \Checkmark & \Checkmark & \Checkmark & \textbf{52.08} & \textbf{82.76} & \textbf{48.34} & \textbf{75.57} \\
     \bottomrule
    \end{tabular}
    \caption{Top-1 accuracy and Clustering results for our method ablations on IN-100 and LV17. We conduct ablations on our discriminative prompt augmentation (Prompt), self-training stage (S.T.), and structural semantic alignment loss ($L_{str}$).}
    \label{tab:method-ablations}
\end{table}

\subsection{Evaluation}
\label{exp:eval}
\subsubsection{Datasets.} 
We evaluate \textbf{S\textsuperscript{3}A} on two generic and five fine-grained benchmarks, \ie, the generic benchmarks of sampled ImageNet-100 (IN100) and ImageNet-1K (IN1K)~\cite{imagenet1k}, and fine-grained benchmarks of StanfordDogs (SDogs)~\cite{dataset:sdogs}, Living17 (LV17), Nonliving26 (NL26), Entity13 (ET13), and Entity30 (ET30) in BREEDS~\cite{dataset:breeds}). 
Furthermore, we evaluate our \textbf{S$^3$A} on three benchmarks for the out-of-vocabulary evaluation (containing categories out of our vocabulary), \ie, Oxford-IIIT Pet (Pet)~\cite{dataset:pet}, CIFAR100~\cite{dataset:cifar100}, and Caltech101(Clatech)~\cite{dataset:caltech}. 
The profile of datasets is listed in our Appendix A. \par 

\subsubsection{Metrics.}
We adopt the top-1 classification accuracy and clustering accuracy (following SCD~\cite{scd} and defined below) for the evaluation.
\begin{equation}
    \textsc{Acc}_{clu}=
    \frac{1}{N} \sum_{i=0}^N{
    \max_{\rho}{\mathbb{I}(y_i=\rho(\hat{y_i}))}
    },
    \label{eq:cluster-acc}
\end{equation}
where $\rho$ is a permutation assignment of cluster indices. $y_i$ and $\hat{y}_i$ are ground-truth predicted categories.
Meanwhile, we adopt Intersection-over-Union (IoU) score as an auxiliary metric in ablations to inspect the overlap between our predictions $\mathcal{Y}_{pred}$ and the ground-truth label set $\mathcal{Y}_{gt}$, \ie, $\frac{|\mathcal{Y}_{pred}\bigcap \mathcal{Y}_{gt}|}{|\mathcal{Y}_{pred}\bigcup \mathcal{Y}_{gt}|}$.
In the out-of-vocabulary experiments, some class names cannot be found in the vocabulary. Thus, we instead apply a soft accuracy score, defined as the similarity between the predicted word (in vocabulary) and the ground truth label. Inspired by BertScore~\cite{bertscore}, we adopt a language model, Sentence-Bert~\cite{sentence-bert}, to calculate the similarity. \par

\subsubsection{Baselines.}
RZSC is a new setting in which few baselines are ready-to-use. Thus, we evaluate the baseline methods by reproducing them with officially released codes in our setting. Specifically, we consider CLIP as the naive baseline, and two state-of-the-art methods in ZST and UF, \ie, SCD~\cite{scd} and MUST~\cite{must}.
In summary, the following baselines are included for performance comparisons:
\begin{itemize}
    \item \textbf{DINO+KMeans}~\cite{dino}: DINO is an contrastive self-supervised learning method. We include it here for clustering quality comparisons. We only report its clustering accuracy as it cannot classify.
    \item \textbf{CLIP}~\cite{clip}: a large-scale VLM pre-trained with massive image-caption pairs conducts zero-shot prediction given our vocabulary.
    \item \textbf{CLIP (Group)}~\cite{clip}: 
    We sequentially conduct clustering, voting, and Hungarian matching on CLIP image embeddings for structural zero-shot transfer, using \textbf{S\textsuperscript{3}A} vocabulary.
    \item \textbf{CLIP (Ideal)}~\cite{clip}: it denotes zero-shot transfer with pre-trained CLIP but given an ideal vocabulary, showcasing the upper bound performance of CLIP representation. 
    \item \textbf{MUST}~\cite{must}: it is an unsupervised ZSC method leveraging instance-wise unsupervised self-training jointly with self-supervised masked-image prediction. We adapt it with our huge vocabulary.
    \item \textbf{SCD}~\cite{scd}: it is an unsupervised/semi-supervised zero-shot transfer method with WordNet vocabulary.
    Its iterative algorithm aligns each cluster with one category name. 
    We adapt it with our \textbf{S\textsuperscript{3}A} vocabulary. During the inference, we classify images with $K$ category prototypes discovered during the training phase.
\end{itemize}

\subsection{Main Results}
To validate the effectiveness of our proposed method, \textbf{S$^3$A}, we conducted an extensive evaluation under RZSC setting. We compared our \textbf{S$^3$A} with various baselines on both fine-grained and generic datasets. The results are in Table~\ref{tab:main-merged}. \par 

Our method, \textbf{S\textsuperscript{3}A}, consistently achieves SOTA results, outperforming CLIP by a substantial margin—specifically, an over $+15\%$ in top-1 accuracy. Furthermore, \textbf{S\textsuperscript{3}A} notably excels over our adapted SOTA baselines, with nearly $+5\%$ in top-1 accuracy and $+6\%$ in clustering accuracy. Generally, we can observe that more classes introduce challenges, and fine-grainedness decreases clustering quality but improves alignment accuracy, \eg, IN100, NL26.
Besides, despite CLIP (Group) being augmented with the same clustering information as \textbf{S\textsuperscript{3}A}, it encounters alignment issues when working with large vocabularies under low-quality clustering, as seen on IN1K and SDogs. We posit this superior performance of \textbf{S\textsuperscript{3}A} may be attributed to its capacity to dynamically calibrate noisy clustering during the self-learning process. It is noteworthy that the existing SOTA in unsupervised zero-shot classification, MUST, at times fails to improve its initial representation when using our realistic vocabulary. This underlines the suboptimality of naive self-training methods for RZSC. \par 

\subsection{Ablations and Analysis}

\para{Method Ablations.} 
To validate the contribution of \textbf{S$^3$A} components, we conduct method ablations on one generic and fine-grained dataset, \ie, IN100 and LV17.
We present the results in Table~\ref{tab:method-ablations}. The last row represents our full method.
When we only keep the initial iterative clustering in our CVPR (the 1$^{st}$ row), our method is equivalent to SCD~\cite{scd}.
The 2$^{nd}$ row denotes our CVPR without all self-training-related components; while, the 3$^{rd}$ row conducts self-training only with instance-wise semantic alignment, similar to MUST~\cite{must}.
The 4$^{th}$ row indicates our \textbf{S\textsuperscript{3}A} without LLM knowledge guidance. 
Based on the results, we can conclude that: \texttt{(1)} Comparing Row 4\&5, although the clustering quality remains comparable without our discriminative prompt augmentation, the semantic alignment degrades, as witnessed by the drop in top-1 accuracy.
\texttt{(2)} Comparing Row 1\&2\&3, self-training with structural alignment dominates the contribution in representation adaptation, witnessed by the cluster performance boosts.
\texttt{(3)} Comparing Row 3\&4, we observe that the structural alignment w/o prompt augmentation yields great improvements on generic datasets, while its effect is less pronounced on fine-grained datasets due to the lack of language signals to discriminate among similar visual categories.
In summary, all components of our \textbf{S$^3$A} enhance the performance. \par

\para{Performance on Estimated $K$.} 
The clustering algorithm requires the number of classes $K$. In Table~\ref{tab:main-merged}, we used the ground truth $K$. In addition, we also implement an iterative algorithm using pre-trained CLIP image features to determine the appropriate $K$, based on the semantic structure of data represented by CLIP~\cite{clip-gcd, scd}, where the detailed algorithm is shown in Appendix C.
We present our experimental results on the estimated $K$ in Table~\ref{tab:uk} on four fine-grained datasets. 
The estimated $K$ are listed beside the dataset title, which is close to the ground-truth values. We show that the estimated $K$ can also bring comparable results with ground truth ones. 
Beyond this experiment, we also perform ablation studies to assess the impact of various estimation error scales on $K$. Detailed results and discussions are in our Appendix D. \par

\begin{table}[t]
    \centering
    \fontsize{9pt}{9pt}
    \selectfont
    \setlength{\tabcolsep}{1.5pt}
    \begin{tabular}{l|cccc}
    \toprule
        Methods & LV17 (73) & NL26 (101) & ET13 (252) & ET30 (206) \\
        \midrule \midrule
        DINO+KMeans & -/72.68 & -/62.93 & -/67.41 & -/63.22 \\
        CLIP & 37.31/47.24 & 33.35/38.96 & 30.09/40.00 & 31.23/39.90 \\
        MUST & 31.71/49.35 & 35.30/48.68 & 38.46/58.25 & 33.41/47.08 \\
        SCD & 40.70/69.17 & 52.63/70.21 & 40.12/71.37 & 45.03/69.14 \\
        \midrule
        \textbf{S$^3$A} (Est-K) & \bf{49.83}/\bf{76.23} & \textbf{57.10}/75.66 & \bf{45.54}/\bf{77.23} & {47.86}/72.75 \\
        \textbf{S$^3$A} (GT-K) & 48.34/75.57 & {56.20}/\bf{75.97} & 45.21/76.92 & \bf{50.41}/\bf{76.14} \\
        \bottomrule
    \end{tabular}
        \caption{Transductive evaluation on four fine-grained benchmarks with estimated cluster numbers (Acc/Cluster). 
         The estimated number is behind the dataset title.}
    \label{tab:uk}
\end{table}
\begin{table}[t]
    \centering
    \fontsize{9pt}{9pt}
    \selectfont
    \setlength{\tabcolsep}{4pt}
    \begin{tabular}{l|ccc}
    \toprule
        Methods & Caltech (0.34) & CIFAR100 (0.12) & Pet (0.62) \\
        \midrule \midrule
        CLIP (Ideal) & 91.25/90.96 & 81.54/81.12 & 90.87/92.39 \\
        \midrule
        CLIP & 50.59/49.66 & 41.62/41.65 & 55.60/57.96 \\
        MUST & 51.20/50.80 & 42.93/42.96 & 58.32/55.83 \\
        SCD & 54.08/54.46 & 42.62/41.64 & 58.57/57.58 \\
        \midrule
        \textbf{S$^3$A} (Our) & \bf{55.29}/\bf{55.55} & \bf{46.10}/\bf{46.40} & \bf{59.00}/\bf{60.57} \\
        \bottomrule
    \end{tabular}
    \caption{Tranductive and inductive evaluation on out-of-vocabulary benchmarks (Train/Test Acc). The OOV ratios for each dataset are provided alongside their respective names. Performance is reported by cosine similarity of generic pre-trained Sentence-BERT, upscaled $\times 100$.}
    \label{tab:oov}
\end{table}

\begin{table}[t]
    \centering
    \fontsize{9pt}{9pt}
    \selectfont
    \setlength{\tabcolsep}{3pt}
    \begin{tabular}{c|l|ccc}
    \toprule
        \#Row & Methods & IN1K & ET13 & ET30 \\
        \midrule \midrule
        1 & CLIP (Ideal) & 65.38/96.58 & 78.43/99.61 & 77.99/99.58 \\
        \midrule
        2 & CLIP & 32.21/\textbf{96.49} & 30.09/\textbf{97.31} & 31.23/\textbf{95.83} \\
        3 & SCD & 37.06/35.09 & 40.46/32.31 & 46.29/39.94 \\
        4 & CHiLS$^*$ & 36.23/34.46 & 41.13/33.33 & 46.09/39.94  \\
        \midrule
        5 & Our (WordNet) & 18.69/18.42 & 21.82/16.85 & 20.38/15.94 \\
        6 & Our (Single) & 37.40/35.74 & 41.13/33.00 & 47.09/40.76 \\
        7 & Our (ChatGPT) & 37.69/36.11 & 42.65/36.84 & 47.43/41.18  \\
        8 & Our (GPT-4) & \textbf{37.95}/36.48 & \textbf{44.98}/37.56 & \textbf{48.37}/42.01 \\
        \bottomrule
    \end{tabular}
    \caption{Ablations on prompt augmentation techniques (Acc/IoU). Performance is reported by cosine similarity of generic pre-trained Sentence-BERT, upscaled $\times 100$.}
    \label{tab:prompt}
\end{table}

\begin{figure}[t]
    \centering
    \includegraphics[width=\linewidth]{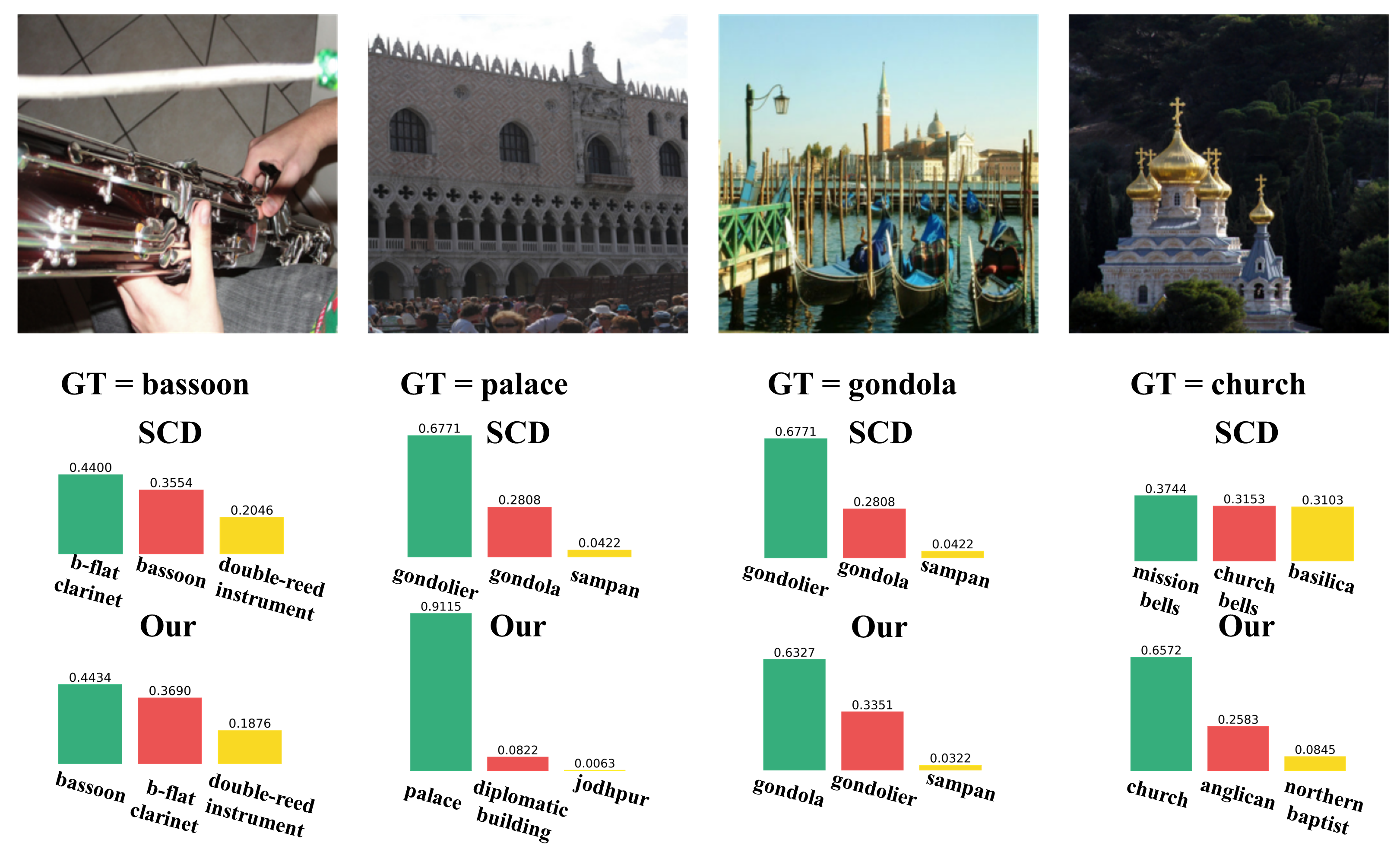}
    \caption{Qualitative results in IN100 without finetuning (SCD~\cite{scd} and our CVPR).}
    \label{fig:example}
\end{figure}

\para{On Out-Of-Vocabulary (OOV) Scenarios.}
Considering the scenarios in which target datasets have category names out of our \textbf{S$^3$A} vocabulary, we further conduct an out-of-vocabulary evaluation on three benchmarks, \ie, Caltech101~\cite{dataset:caltech}, CIFAR100~\cite{dataset:cifar100}, and Oxford-IIIT Pet~\cite{dataset:pet}. 
The out-of-vocabulary ratios of datasets and results are presented in Table~\ref{tab:oov}.
We can conclude that \textbf{S$^3$A} still achieves SOTA performance in this challenging setup on both inductive and transductive evaluation. \par

\para{On Effectiveness of \textbf{S$^3$A} Prompt Augmentation.}
In this ablation experiment, we analyze the effect of the proposed LLM-guided discriminative prompt augmentation in our CVPR algorithm.
We compare with four augmentation setups in Table~\ref{tab:prompt}: \texttt{(1)} using WordNet definition for augmentation (5$^{th}$ row); \texttt{(2)} reduce prompt semantic discriminativeness by requesting visual attributes for only a single category name in each LLM prompt (6$^{th}$ row); \texttt{(3)} our prompt augmentation guided by ChatGPT (7$^{th}$ row); \texttt{(4)} our prompt augmentation guided by GPT-4.
Besides, we also compare with a recent SOTA, CHiLS~\cite{chils}, in prompt augmentation for zero-shot prediction. We use their prompt to generate ten subcategories for each class.
We can draw the following conclusions: \texttt{(1)} Semantic distinctiveness in prompts aids fine-grained differentiation; \texttt{(2)} Incorporating WordNet linguistic knowledge hinders semantic discriminativeness; 
\texttt{(3)} Our approach outperforms CHiLS, thus is more tailored to RZSC tasks; 
\texttt{(4)} CLIP focuses on instance alignment and leads to low ACC but high IoU; 
\texttt{(5)} Our method benefits from advanced LLMs. \par

\para{Qualitative Examples}
We present qualitative examples from IN100 in Fig.~\ref{fig:example}, in which four sampled cluster-level predictions of SCD and \method-CVPR on top-3 categories are displayed with corresponding exemplar images, to illustrate the superiority of our semantic structural alignment.
The results demonstrate that our offline CVPR algorithm can effectively correct category misrecognitions and precisely focus on salient categories. \par

\section{Conclusion}
In this work, we address the challenging task of Realistic Zero-Shot Classification, without assuming partial source supervision or ideal vocabularies. We propose a Self Structural Semantic Alignment (\textbf{S$^3$A}) framework, anchored by an innovative Cluster-Vote-Prompt-Realign (CVPR) algorithm for structural semantic relationship mining and a self-training process for iterative semantic alignment. Our experiments demonstrate the effectiveness of \textbf{S$^3$A}, achieving significant accuracy improvements over baseline methods on all generic and fine-grained benchmarks. Our results emphasize the cooperation between the self-training and our CVPR structural alignment approach. Our approach exhibits superiority over other prompt augmentation strategies, with unknown class numbers, and in out-of-vocabulary scenarios. \par

\bibliography{aaai24}

\appendix
\renewcommand{\thepage}{S-\arabic{page}}
\renewcommand{\thesection}{S-\arabic{section}}
\renewcommand{\thetable}{S-\arabic{table}}
\renewcommand{\thefigure}{S-\arabic{figure}}

\twocolumn[
\begin{@twocolumnfalse}
\section*{\centering {\Large Towards Realistic Zero-Shot Classification via Self Structural Semantic Alignment -- {\em Supplementary Material}}}
\end{@twocolumnfalse}
]

\section{A Dataset Profile}
\begin{table*}[ht]
    \centering
    \fontsize{9pt}{9pt}
    \selectfont
    \begin{tabular}{c|ccccccc}
    \toprule
        Dataset & StanfordDogs & ImageNet-100 & ImageNet-1K & Living17 & Nonliving26 & Entity13 & Entity30 \\
    \midrule
        \#Train & 12K & 127K & 1M & 88K & 132K & 334K & 307K \\
        \#Test & 8K & 5K & 50K & 3K & 5K & 13K & 12K \\
        \#Category & 120 & 100 & 1000 & 68 & 104 & 260 & 240 \\
    \bottomrule
    \end{tabular}
    \caption{\textbf{The profile of seven benchmarks for performance comparisons.} Here, we report the training set size, test set size, and category name.}
    \label{tab:dataset}
\end{table*}

\begin{table}[ht]
    \centering
    \fontsize{9pt}{9pt}
    \selectfont
    \begin{tabular}{c|ccc}
    \toprule
        Dataset & Caltech101 & CIFAR100 & Pet \\
    \midrule
        \#Train & 6K & 50K & 3K\\
        \#Test & 1K & 10K & 3K\\
        \#Category & 100 & 100 & 37\\
        \%OOV & 34 & 12 & 62\\
    \bottomrule
    \end{tabular}
    \caption{\textbf{The profile of three benchmarks for out-of-vocabulary evaluation.} Here, we report the training set size, test set size, total category number, and out-of-vocabulary ratio of category names.}
    \label{tab:dataset2}
\end{table}

We report the details of our benchmarks in our main results in Table~\ref{tab:dataset}. Additionally, we also present the out-of-vocabulary benchmark details in Table~\ref{tab:dataset2}. \par 

\section{B Additional Related Work}
\para{Prompt Augmentation.} 
Prompt augmentation has proven conducive to multi-modal zero-shot transfer by providing supplementary information in addition to raw category names to better capture the training data distribution~\cite{prompt-nlp-survey, clip}.
The simplest practice is ensembling prompts with handcrafted templates~\cite{clip}.
Further, driven by the powerful knowledge retrieval capability of emergent LLMs~\cite{gpt-3, prompt-vqa}, some recent works \cite{chils, customized-prompt, point-clip} propose to augment text-prompt templates by querying LLM for class-specific semantic information (e.g., attributes) or incorporating dataset-specific domain knowledge. 
Based on our empirical comparisons, our proposed VTA augmentation is better at distinguishing between fine-grained semantic categories. \par

\para{Learning with Large Vocabulary.}
Recently, large vocabulary vision tasks grow popular in the research community.
POMP~\cite{pomp} proposes to utilize few-shot examples in ImageNet-21K~\cite{imagenet21k} to enhance the generalization capability of CLIP on downstream tasks/datasets.
Detic~\cite{detic} proposes to employ massive image-labeled images from ImageNet-21K to tune a detector classifier of over 20K categories for open-vocabulary object detection.
Meanwhile, \cite{large-vocab-detection-baseline} also proposes a simple two-stage baseline for the object detection task, which jointly learns from large-vocabulary classification data and small-vocabulary detection.
However, their methods are orthogonal to ours, since they treat images from a large number of categories as labeled source data to benefit generalization towards downstream datasets; while our learning goal is to categorize an unseen unlabeled dataset with no supervision. \par

\section{C S$^3$A Training Algorithm}
\subsection{More Details on CVPR Algorithm}
\para{Iterative Clustering.} In the clustering step of our CVPR, we incorporate additional iterative clustering steps~\cite{scd}, which refine the previous cluster results with voting results on textual prototypes.
Suppose the clustering partition at $t$-th step is $\Gamma^t$, and its top-1 voting matrix is $M^t \in \mathcal{R}^{K \times |\mathcal{W}|}$, we assign each cluster with a distinct category name by Hungarian matching, \ie, solving maximum-weight bipartite matching between $K$ clusters and $|\mathcal{W}|$ words based on the weight matrix $M^t$.
Formally, this optimization problem of bipartite matching is formulated as:
\begin{equation}
\begin{array}{llcl}
\max_{A} & tr(M^t A^T) & & \\
\text{s.t.} & \displaystyle\sum_{j=1}^{|\mathcal{W}|} A_{kj} = 1, & \forall k \in \{1, \ldots, K\} & \\
& \displaystyle\sum_{k=1}^{K} A_{kj} \leq 1, & \forall j \in \{1, \ldots, |\mathcal{W}|\} & \\
& A_{kj} \in \{0, 1\}, & \forall k,j &
\end{array}
\label{eq:hungarian}
\end{equation}
where $A \in \mathcal{R}^{K \times \mathcal{|W|}}$ is the 0-1 assignment matrix. Here, $A_{k,j}=1$ denotes the $j$-th word is assigned to the $k$-th cluster, and $A_{k,j}=0$ otherwise. We omit the superscript $t$ for $A$ for notation simplicity. \par 

After $A$ has been resolved, the textual prototypes of $K$ assigned words are considered as common argmax (pseudo) classifiers which cluster all image embeddings into the updated partition $\Gamma^{t+1}$. Then, the image clusters $\Gamma^{t+1}$ follow the same previous process. 
This iteration terminates until the partition $\Gamma^{t+1}$ does not change any further. \par

\begin{table}[]
    \centering
    \begin{tabular}{c|cccc}
    \toprule
        Method & LV17 & NL26 & ET13 & ET30 \\
    \midrule
        Ground-truth & 68 & 104 & 260 & 240 \\
        Single-pass & 52 & 87 & 323 & 201 \\
        Iterative (Our) & 73 & 101 & 252 & 206 \\
    \bottomrule
    \end{tabular}
    \caption{Different methods to estimate $K$.}
    \label{tab:estimation-of-cluster}
\end{table}

\para{Estimation of $K$ using CLIP.} We propose a simple iterative estimation technique to estimate the class number of an unlabeled dataset based on CLIP image embeddings. 
GCD~\cite{gcd} utilizes the elbow algorithm to determine the optimal cluster count $K$ by pinpointing the inflection in cluster scores across a predefined $K$ range. This 'elbow' or inflection arises when there is a noticeable deceleration in the reduction of clustering scores. 
For our clustering metric, we employ the Silhouette score~\cite{silhouette}. 
Relying solely on a single-pass elbow algorithm can make the estimated $K$ vulnerable to noise. To address this, our iterative approach consists of three passes: first, we scan the range [$\text{LB}_0$, $\text{UB}_0$], then [$\text{LB}_0$, $\text{S}_1$], and finally [$\text{S}_2$, $\frac{\text{S}_2 + \text{S}_1}{2}$], each time applying the elbow algorithm. Here, $\text{S}_1$ and $\text{S}_2$ denote the solution of the first and second pass.
In practice, we consistently set $\text{LB}_0=50$ and $\text{UB}_0=2000$ across all datasets, and our experiments show that the iterative method offers improved precision over the single-pass approach (see Table~\ref{tab:estimation-of-cluster}). \par 

\begin{table}[t]
    \centering
    \fontsize{9pt}{9pt}
    \selectfont
    \begin{tabular}{c|ccccc}
    \toprule
        Params & IN/BREEDS & SDogs & Caltech & CIFAR & Pet \\
        \midrule
        Init. EMA & 0.999 & 0.999 & 0.999 & 0.999 & 0.99 \\
        EMA Iter. & 2000 & 2000 & 2000 & 500 & 100 \\
        $\tau$ & 0.5 & 0.5 & 0.5 & 0.3 & 0.7 \\
        Epoch & 30 & 30 & 30 & 30 & 60 \\
     \bottomrule
    \end{tabular}
    \caption{\textbf{Hyperparameters for all benchmarks}, including the initial EMA weight decay value, EMA warmup iterations, confidence parameter $\tau$, and training epoch number.}
    \label{tab:hyperparam}
\end{table}

\begin{table}[tp]
    \centering
    \fontsize{9pt}{9pt}
    \selectfont
    \setlength{\tabcolsep}{4pt}
    \begin{tabular}{l|ccccc|c}
    \toprule
        Methods & IN100 & LV17 & NL26 & ET13 & ET30 & Avg \\
         \midrule
         CLIP (Ideal) & 89.09 & 80.17 & 84.38 & 83.87 & 82.33 & 83.77 \\
         \midrule
         CLIP & 34.82 & 43.59 & 36.95 & 35.62 & 36.44 & 37.48 \\
         SCD & 50.61 & 54.10 & 55.23 & 48.82 & 49.44 & 51.64 \\
         \midrule
         \textbf{S$^3$A}-CVPR (Our) & \textbf{52.51} & \textbf{55.31} & \textbf{56.02} & \textbf{51.31} & \textbf{52.78} & \textbf{53.59} \\
         \bottomrule
    \end{tabular}
    \caption{\textbf{Transductive evaluation on five benchmarks for backbone ablations.} Top-1 classification accuracy scores are reported in percentage. We highlight the highest scores except for the upper bound. The evaluation is conducted for the CLIP ViT-L/14 backbone. We present results from our \textbf{S$^3$A}-CVPR without training.}
    \label{tab:backbone}
\end{table}

\para{Training Algorithm.}
In this part, we delineate our \textbf{S$^3$A} training algorithm, which is presented in algo.~\ref{algo}.
At each training epoch, we first conduct our CVPR algorithm on all extracted image embeddings from the teacher model to obtain the semantic structural alignment labels $\mathbf{\hat{Y}}$ (from line 6 to line 18). During the training (from line 19 to line 24), we compute two loss functions: \texttt{(1)} instance semantic alignment loss between one-shot pseudo-label from the teacher model, \ie, the nearest neighbor prediction of each instance on our vocabulary; \texttt{(2)} structural semantic alignment loss between student predictions and our structural pseudo labels $\mathbf{\hat{Y}}$. 
Finally, the teacher is EMA updated with EMA decay $\eta$ by student parameters. \par

\begin{figure*}[t]
    \centering
    \includegraphics[width=\linewidth]{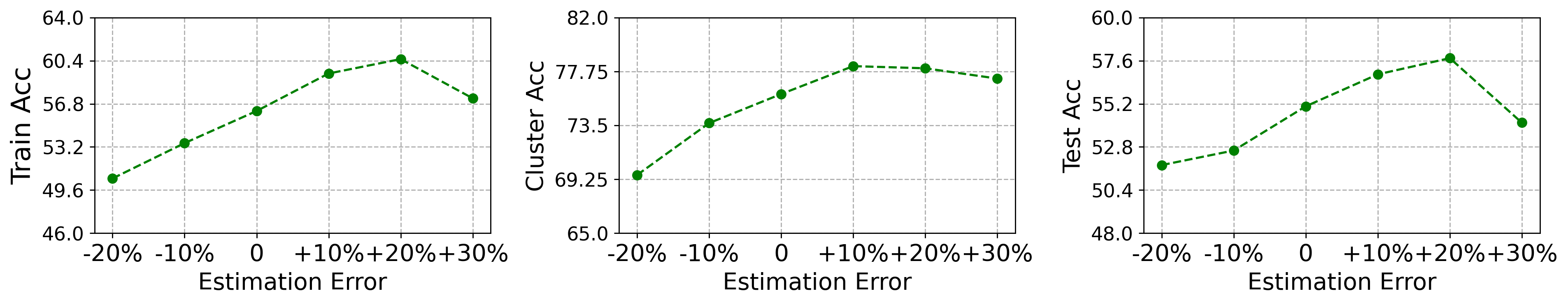}
    \caption{\textbf{Assessing sensitivity to errors in $K$ estimation.} We present results corresponding to different percentages of estimation inaccuracies relative to the true value of $K$.}
    \label{fig:estimation-error}
\end{figure*}

\begin{figure*}[t]
    \centering
    \includegraphics[width=\linewidth]{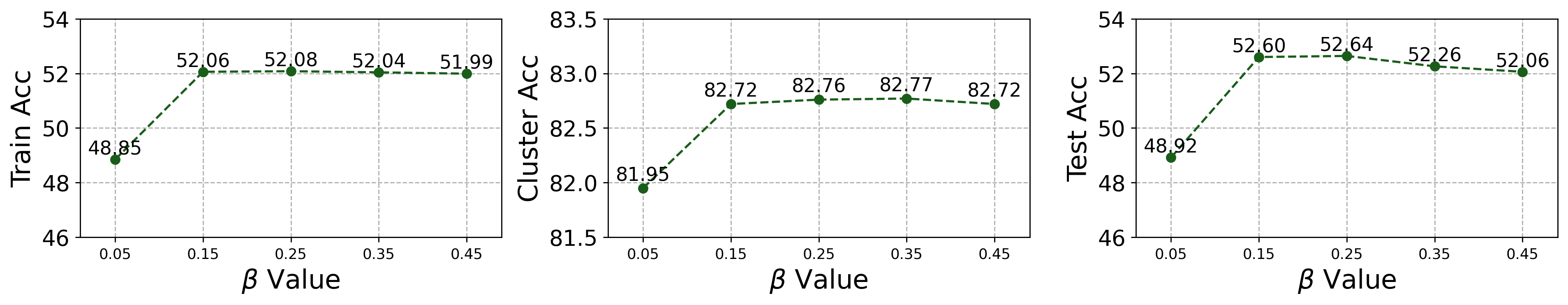}
    \caption{\textbf{Ablation study on the structural semantic alignment loss weight $\beta$.} By default, we set $\beta=0.25$.}
    \label{fig:beta}
\end{figure*}

\subsection{Implementation Details of S$^3$A}
For our method, we construct our \textbf{S$^3$A} huge taxonomic vocabulary from all distinct class names in ImageNet-21K~\cite{imagenet21k} and ImageNet-1K~\cite{imagenet} datasets, which is comprehensive enough to cover all common classes in downstream datasets~\cite{concept-gen, imagenet21k}.
We fix $m=3$ and $\gamma=0.25$ across all datasets. 
The class number $K$ of the target dataset is assumed known unless specified. Considering requesting LLM is time-demanding, we only compute prompting at the first epoch. \par 

We apply strong augmentations for the student, \ie, RandomResizedCrop, RandomFlip, and RandAug~\cite{must-36}, and weak augmentations for the teacher, \ie, Resize, and RandomCrop.
We adopt ViT-B/16~\cite{vit} as our CLIP backbone for main evaluation.
Following \cite{must}, we utilize AdamW~\cite{must-34} optimizer with a learning rate of $1e^{-5}$, a batch size of $128$, and a weight decay of $0.05$. The cosine learning rate schedule without warmup is adopted.
Following \cite{must-5}, we also adopt layer-wise learning rate decay of $0.65$. 
Notice that we train for 30K iterations at maximum across all datasets.
Besides, we present details on other hyperparameters in Table~\ref{tab:hyperparam}.
Specifically, we linearly warmup the EMA decay parameter from the specified initial EMA value to $0.9998$ within specified iterations in Table~\ref{tab:hyperparam}, following \cite{must}.
We observe negligible random variances in final results since our \textbf{S$^3$A} is deterministic and our randomness only comes from the optimizer. 
For backbone ablations, we increase the initial learning rate to $2e^{-5}$ for ViT-L/14 backbone.
During the inference, we adopt the teacher model for inference with the weak data augmentation on our entire \textbf{S$^3$A} vocabulary for predictions. 
All experiments are conducted on a single RTX A6000 GPU. \par

\section{D Additional Experimental Results}

\begin{table}[tp]
    \centering
    \fontsize{9pt}{9pt}
    \selectfont
    \setlength{\tabcolsep}{4pt}
    \begin{tabular}{l|ccc}
    \toprule
        Methods & Train Acc & Cluster Acc & Test Acc \\
         \midrule
         CLIP (Ideal) & 83.87 & 83.21 & 83.20 \\
         \midrule
         CLIP (ViT-L/14) & 35.62 & 40.00 & 33.91 \\
         SCD (ViT-L/14) & 48.82 & 76.24 & 46.91 \\
         MUST (ViT-L/14) & 43.29 & 58.25 & 37.63 \\
         \midrule
         \textbf{S$^3$A} (ViT-B/16) & 45.21 & 76.92 & 45.65 \\
         \textbf{S$^3$A} (ViT-L/14) & \textbf{52.61} & \textbf{81.00} & \textbf{53.32} \\
         \bottomrule
    \end{tabular}
    \caption{\textbf{Transductive evaluation on BREEDS-ET13 benchmarks for backbone ablations.} We highlight the highest scores except for the upper bound. The evaluation is conducted for the CLIP ViT-L/14 backbone.}
    \label{tab:backbone-2}
\end{table}

\begin{table}[t]
    \centering
    \fontsize{9pt}{9pt}
    \selectfont
    \begin{tabular}{c|ccccc}
    \toprule
        Method & SDogs & IN100 & ET13 & ET30 & Pet \\
    \midrule
        BLIP-2 & 43.49 & 47.57 & 54.60 & 52.96 & 54.05\\
        BLIP-2 (Group) & 45.01 & 51.65 & 59.60 & 58.17 & 55.12\\
    \midrule
        \textbf{S$^3$A} (Our) & \textbf{78.45} & \textbf{75.77} & \textbf{70.73} & \textbf{73.86} & \textbf{59.00} \\
    \bottomrule
    \end{tabular}
    \caption{\textbf{Additional evaluation for comparing with BLIP-2 LMM.} The scores are reported with soft accuracies based on Sentence-BERT similarity.}
    \label{tab:compare-lmm}
\end{table}

\para{Ablations on Different Backbones.}
Besides the main results in Table~2, we conduct additional ablations on our offline CVPR algorithm with different CLIP backbones, \ie, ViT-L/14, and report our results in Table~\ref{tab:backbone}.
We can observe that, even without self-alignment fine-tuning, our method can considerably surpass our adapted SOTA, \ie, SCD, with nearly $2\%$ in the top-1 accuracy.
Moreover, we present our results in Table~\ref{tab:backbone-2} on a randomly chosen benchmark, \ie, BREEDS-Entity13, to compare our entire method with other baselines.
Further, we conclude that our entire method with self semantic alignment tuning can achieve significant performance advancement with $\sim 4\%$ in classification and $\sim 5\%$ in clustering w.r.t. existing SOTA methods. \par 

\para{Generalization under Inductive Evaluation.} Apart from transductive scenarios, we evaluate the generalizability of \method under inductive setting on across all benchmarks. The results are presented in Table~\ref{tab:main-induct}, which demonstrates that \method can generalize to unseen test images with comparable performance on the unlabeled training data. \par

\begin{table}[t!]
    \centering
    \fontsize{9pt}{9pt}
    \selectfont
    \setlength{\tabcolsep}{4pt}
    \begin{tabular}{l|ccccccc}
    \toprule
        Methods & SDogs & IN100 & IN1K & LV17 & NL26 & ET13 & ET30 \\
         \midrule
         MUST & 58.88 & 33.90 & 28.72 & 31.36 & 33.85 & 37.57 & 33.03 \\
         \textbf{S$^3$A} (Ind.) & \textbf{61.31} & \textbf{52.64} & \textbf{42.61} & \textbf{47.56} & \textbf{55.06} & \textbf{45.65} & \textbf{50.12} \\
         \midrule
         \textbf{S$^3$A} (Trs.) & 58.94 & 52.08 & 42.43 & 48.34 & 56.20 & 45.21 & 50.41 \\
         \bottomrule
    \end{tabular}
    \caption{\textbf{Inductive evaluation on seven benchmarks.} Top-1 classification accuracy scores on the test set are reported in percentage. Here, \textbf{S$^3$A} (Ind.) and \textbf{S$^3$A} (Trs.) denotes \method under inductive/transductive evaluation.}
    \label{tab:main-induct}
\end{table}

\para{Sensitivity to $K$.}
To determine the underlying effects of estimation error on $K$ on our \textbf{S$^3$A} method, we further assess the sensitivity of \textbf{S$^3$A} performance w.r.t. this estimation error. 
In Fig.~\ref{fig:estimation-error}, we showcase results for values of $K$ varying from a decrease of $20\%$ to an increase of $30\%$ from its true value. 
From our analysis, it is evident that our \textbf{S$^3$A} demonstrates substantial resilience to over-fragmentation. However, with an equivalent degree of under-fragmentation error, there is a noticeable performance drop due to a larger increase in the number of instances inaccurately assigned. \par

\para{Sensitivity to $\beta$.}
We conduct ablation study on the robustness of our \textbf{S$^3$A} towards the weight $\beta$ of the structural semantic alignment loss. The results are presented in Table~\ref{fig:beta}.
In this experiment, we vary the $\beta$ by $0.1$ and $0.2$ respectively on both sides.
From the findings, we deduce that our \textbf{S$^3$A} remains robust with variations in $\beta$, unless $\beta$ is exceedingly small. As $\beta$ increases, there's a marginal decline in overall performance. \par

\para{Additional Comparisons with the Large-Multimodal Model (LMM).}
We further present additional results for comparisons between our \textbf{S$^3$A} and the recently proposed powerful LMM, \ie, BLIP-2~\cite{blip-2}, on several benchmarks to showcase the effectiveness of our method. 
To ensure fair comparisons, we adopt pre-trained BLIP-2 with ViT-B/16 backbone and reproduce the method based on their official code.
Our study includes two baselines: \texttt{(1)} In the naive CLIP-2 baseline, we prompt with ``What is the category name of the object in the photo? {$c_0$}, {$c_1$}, or {$c_2$}?'' (where $c_0, c_1, c_2$ are candidate outputs from SCD~\cite{scd}). Here, we treat the generated names as predictions.
\texttt{(2)} For the CLIP-2 (Group) baseline, the category name with the highest voting frequency in each cluster (determined using KMeans~\cite{kmeans} clustering algorithm) is designated as the prediction for all images within that specific cluster.
Given that the LMM output can sometimes fall outside of the expected vocabulary, we utilize soft accuracy scores for the evaluation.
The experimental results are displayed in Table~\ref{tab:compare-lmm}. We can conclude that our \textbf{S$^3$A} consistently outperforms the powerful BLIP-2~\cite{blip-2}, showcasing its superiority in handling generic, fine-grained, and out-of-vocabulary benchmarks. 
We speculate that the low performance of BLIP-2 might originate from its unpredictable text generation tendencies evidenced by its occasional production of unrelated words. \par

\begin{figure*}
    \centering
    \includegraphics[width=\linewidth]{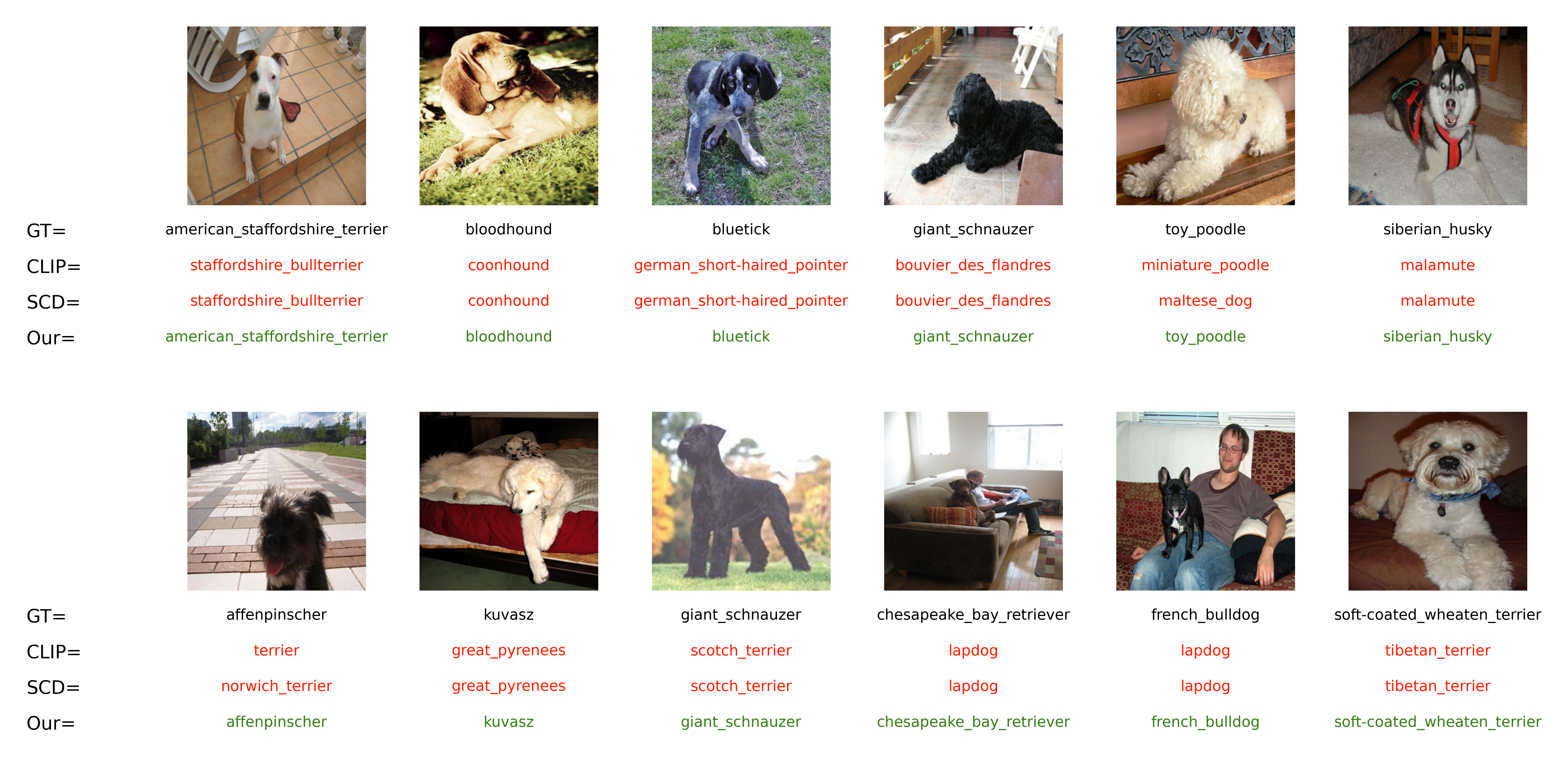}
    \caption{\textbf{Additional qualitative examples on StanfordDogs dataset.} The ground-truths, and category names predicted by CLIP, SCD, and our \textbf{S$^3$A} are presented below each image. Green/red texts denote correct/incorrect predictions.}
    \label{fig:visualize-sdogs}
\end{figure*}

\begin{figure*}
    \centering
    \includegraphics[width=\linewidth]{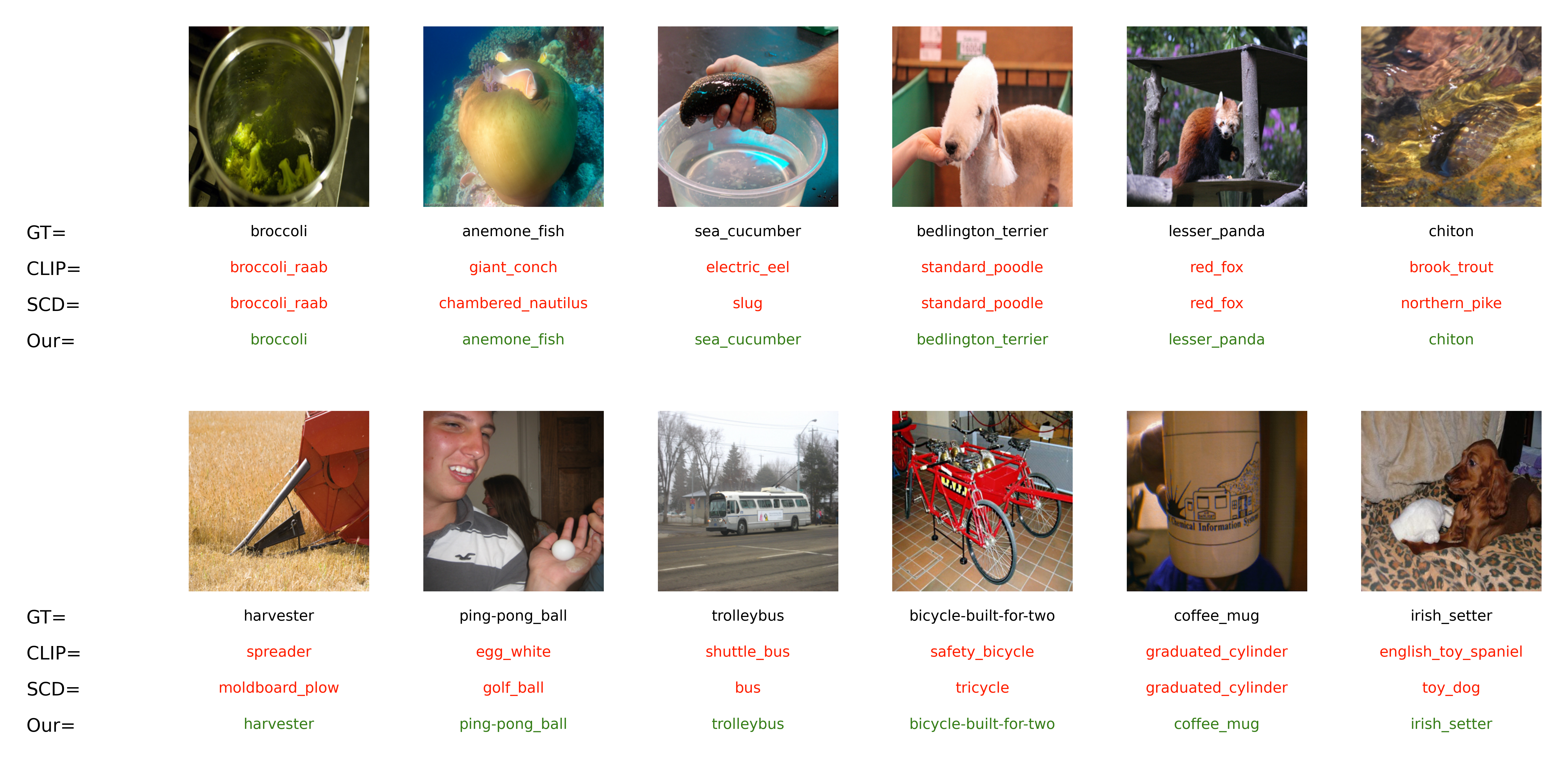}
    \caption{\textbf{Additional qualitative examples on ImageNet-1K dataset.} The ground-truths, and category names predicted by CLIP, SCD, and our \textbf{S$^3$A} are presented below each image. Green/red texts denote correct/incorrect predictions.}
    \label{fig:visualize-imagenet1k}
\end{figure*}

\para{Additional Qualitative Examples.}
We present additional qualitative examples in Fig.~\ref{fig:visualize-sdogs} and Fig.~\ref{fig:visualize-imagenet1k} on SDogs and IN1K to evaluate our \textbf{S$^3$A} method against other baselines.
Our method is robust to both textual fine-grained category names (\eg, ``trolleybus'' and ``shuttle bus'') and visual fine-grainedness (\eg, ``bloodhood'' and ``coonhood'', ``ping-pong ball'' and ``egg white'' or ``gulf ball''), which corroborates our \textbf{S$^3$A} method has learned more discriminative multi-modal semantic alignments than pre-trained CLIP. \par

\begin{algorithm*}[]
\algsetup{linenosize=\tiny}
\DontPrintSemicolon
\small

  \KwInput{Unlabeled Training dataset $\mathcal{D}_u$, a pre-trained VLM (CLIP) with image encoder $f_I(\cdot | \theta)$ and frozen text encoder $f_T(\cdot)$, large vocabulary $\mathcal{W}$.
  }
  
  \KwOutput{Trained adapted model $f_I(\cdot | \theta)$.
  }

  \text{Initialize teacher encoder $f_{I, T}(\cdot | \theta_S)$ from the student model $f_I$}.\;
  \text{Precompute textual embeddings $\mathbf{H}=f_T(\mathcal{W})$}.\;
  
\For{each epoch $e$=1...$E_1$ }{
    // \textbf{Offline: CVPR Algorithm} \\
    $\mathbf{Z}=\phi, \mathbf{\tilde{Y}}=\phi$\; 
      \For{each batch $\mathbf{X} \in \mathcal{D}_u$}{
          $\mathbf{Z} = \left[\mathbf{Z}; f_{I, T}(\mathbf{X})\right]$ \tcp{collect features}  \par 
          $\mathbf{\tilde{Y}} = \left[\mathbf{\tilde{Y}}; \text{Nearest}(\mathbf{Z}, \mathbf{H})\right]$ \tcp{collect predictions on vocabulary}
      }
      
    // Iterative Clustering \\
    $\Gamma^0=\text{KMeans}(\mathbf{Z}), t=0$ \tcp{initial clustering}
      \For{$t\leq \text{max\_iter}$}{
      Compute $M^t$ from $\Gamma^t$ with eq.~2 ($m=1$) and $\mathbf{\tilde{Y}}$. \tcp{voting}
      Compute cluster-category assignment $A$ from $M^t$ with eq.~\ref{eq:hungarian}. \tcp{Hungarian matching}
      Compute new cluster partition $\Gamma^{t+1}$ by assigning instances $\mathbf{Z}$ to $K$ prototypes from $\{\mathbf{H}_j|\sum_{k=1}^{K}{A_{k, j}}=1\}$.\;
      $t = t + 1$.\;
      }
      
      // Voting \\
      Let $\Gamma=\Gamma^t$. Compute $M$ from $\Gamma$ with eq.~2 ($m=3$) and $\mathbf{\tilde{Y}}$.\;
      Obtain $\{\mathcal{A}_k\}_{k=1}^K$ with prompt augmentation. \tcp{prompting}
      Compute $\tilde{M}$ with eq.~3 and obtain structural semantic alignment $\mathbf{\hat{Y}}$ from Hungarian solution $A$ on $\tilde{M}$. \tcp{realigning}
      // \textbf{Online: Self Training with Semantic Alignment} \\
      \For{each batch $(\mathbf{X}, \mathbf{\hat{Y}}) \in \mathcal{D}_u$}{
          $\mathbf{\tilde{Y}}=\text{Nearest}(f_{I, T}(\mathbf{X}), \mathbf{H})$. \tcp{forward teacher}
          $\mathbf{\tilde{Y}}_S=f_{I}(\mathbf{X})$. \tcp{forward student} 
          // Individual Semantic Alignment Loss \\
          Compute cross entropy between $\mathbf{\tilde{Y}}_S$ and $\mathbf{\tilde{Y}}$ threshold by confidence $\tau$.\;
          // Structural Semantic Alignment Loss \\
        Compute cross entropy between $\mathbf{\tilde{Y}}_S$ and $\mathbf{\hat{Y}}$.\;
          Update teacher parameters $\theta_S = \eta\theta_S + (1 - \eta) \theta$.\;
      }
}      
  \Return{$f_{I, S}(\cdot | \theta_S)$}\;
\caption{\textbf{S$^3$A} Training Algorithm.}
\label{algo}
\end{algorithm*}

\end{document}